\documentclass{article}
\usepackage{spconf,amsmath,graphicx}
\usepackage{amssymb}
\usepackage{amsmath}
\usepackage{algorithm}  
\usepackage{algorithmicx}  
\usepackage{algpseudocode}  
\usepackage{booktabs}
\usepackage{multirow}
\usepackage{colortbl}
\usepackage[table,xcdraw]{xcolor}

\title{TNTC: two-stream network with transformer-based complementarity for gait-based emotion recognition}
\name{Chuanfei Hu$^1$, Weijie Sheng$^1$, Bo Dong$^2$, Xinde Li$^1$}
\address{\small$^1$Key Laboratory of Measurement and Control of CSE Ministry of Education, School of Automation, Southeast University, Nanjing, China \\
	\small $^2$College of Biomedical Engineering \& Instrument Science, Zhejiang University, Hangzhou, China \\
	\small \{cfhu, wjsheng, xindeli\}@seu.edu.cn, bodong.cv@gmail.com
	\thanks{Corresponding author: Xinde Li (xindeli@seu.edu.cn).}
	}
\begin{document}
%
\maketitle
\begin{abstract}
Recognizing the human emotion automatically from visual characteristics plays a vital role in many intelligent applications. 
Recently, gait-based emotion recognition, especially gait skeletons-based characteristic, 
has attracted much attention, while many available methods have been proposed gradually.
The popular pipeline is to first extract affective features from joint skeletons, 
and then aggregate the skeleton joint and affective features as the feature vector for classifying the emotion.
However, the aggregation procedure of these emerged methods might be rigid, 
resulting in insufficiently exploiting the complementary relationship between skeleton joint and affective features.
Meanwhile, the long range dependencies in both spatial and temporal domains of the gait sequence are scarcely considered.
To address these issues, we propose a novel two-stream network with transformer-based complementarity, termed as TNTC. 
Skeleton joint and affective features are encoded into two individual images as the inputs of two streams, respectively.
A new transformer-based complementarity module (TCM) is proposed to bridge the complementarity between two streams hierarchically via capturing long range dependencies.
Experimental results demonstrate TNTC outperforms state-of-the-art methods on the latest dataset in terms of accuracy.

\end{abstract}

\vspace{-2mm}
\section{Introduction} \label{sec:intro}
Human emotion recognition, based on visual cues, has been widely applied in various intelligence applications, 
such as video surveillance \cite{Nasim2021Wearable}, behavior prediction \cite{Montero2021Review, Li2019Fusion}, 
robot navigation \cite{Aniket2019Modelling} and human-machine interaction \cite{Zhang2021Predicting}.
Facial expression is one of the most predominant visual cues \cite{Li2020Deep} to be used for recognizing the human emotions including anger, disgust, happiness, sadness, fear and other combinations.
However, facial expressions may be unreliable in complex situations, for instance, imitation expressions \cite{mondillon2007imitation} and concealed expressions \cite{porter2008reading}.
Therefore, recent researches gradually focus on the other visual cues of human to perceive the emotions, such as a gait of human in a walking \cite{xu2020emotion, sheng2021multi}. 

%

Recent efforts have been made towards improving gait-based emotion recognition \cite{Bhattacharya2020Take, randhavane2019identifying, Bhattacharya2020STEP, zhuang2020g, Narayanan2020ProxEmo}, 
which can be categorized as sequence-based, graph-based and image-based methods.
The paradigm of sequence-based methods is to construct a sequence deep model, 
such as Gate Recurrent Unit (GRU) and Long Short-term Memory (LSTM), based on skeleton sequences to predict the emotion \cite{Bhattacharya2020Take,randhavane2019identifying}.
The insight of graph based-methods is to utilize Spatial Temporal Graph Convolutional Network (ST-GCN) to represent the inherent relationship between joints,
since the skeleton is naturally structured as a graph in non-Euclidean geometric space \cite{Bhattacharya2020STEP, zhuang2020g}.
The image-based methods cast the sequence classification as an image classification via encoding the skeleton sequences,
while Convolutional Neural Network (CNN) is constructed to extract hierarchical features for recognizing the emotions \cite{Narayanan2020ProxEmo}.
Although these methods achieve promising results in human emotion recognition, 
there are two major drawbacks.
Firstly, the aggregation of joints and affective features might be rigid, 
resulting in exploiting the complementary information insufficiently.
Furthermore, long range dependencies in both spatial and temporal domains are ignored,
which are important to depict the implicit relationships between skeleton joints for human poses \cite{Mao2019Learning}.

\begin{figure*}[!t]\centering
	\includegraphics[width=0.8\linewidth]{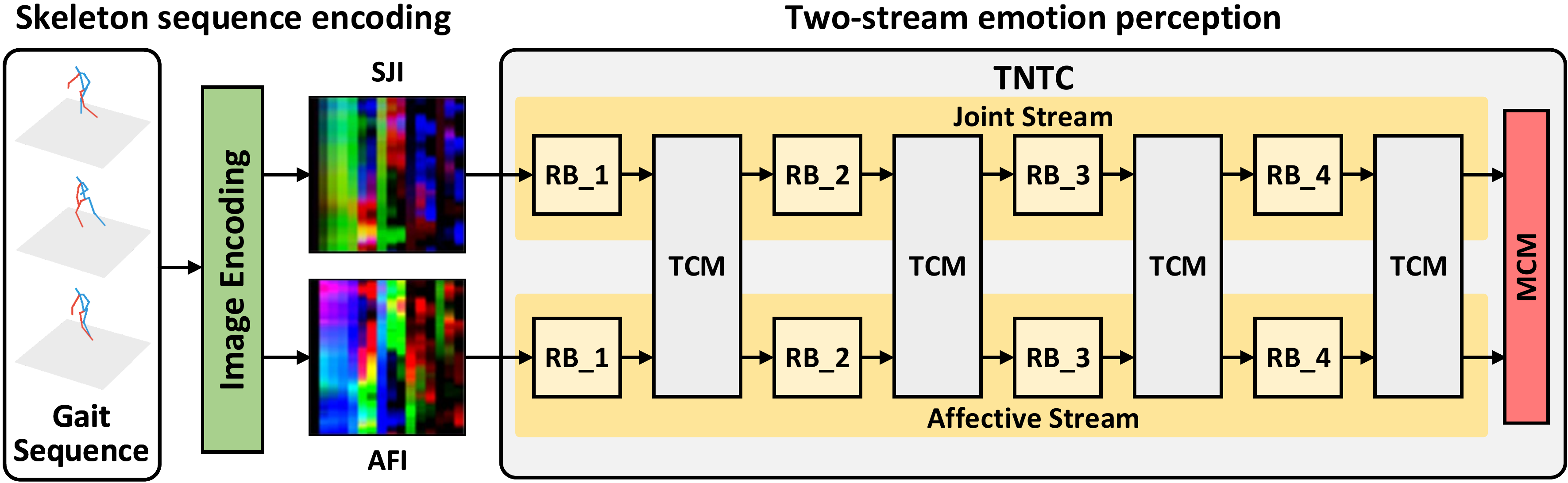}
	\vspace{-3mm}
	\caption{The overall framework of the proposed method. 
		``RB\_id'' denotes the ``id''-th residual block of ResNet-34. 
	}
	\label{fig:framework}
\end{figure*}

To address the above issues, we focus on the image-based method,
and a novel two-stream network with transformer-based complementarity, termed as TNTC, is proposed for recognizing the human emotions.
We argue that spatial and temporal information of the skeleton sequences can be effectively extracted via CNN in the image domain.
Meanwhile, transformers are utilized to handle the problem of capturing long range dependencies via the self-attention mechanism 
whose effectiveness has been verified in many computer vision tasks \cite{dosovitskiy2020image, he2021transreid}.
Specifically, skeleton joints are first encoded into an image,
while affective features are converted to another image via a novel encoded method.
Two individual CNNs are built to extract the hierarchical representations separately, and then
transformer-based complementarity module (TCM) is proposed to
bridge the complementarity between two streams in each feature level via capturing long range dependencies.
The probabilities of human emotions are predicted by the multi-layer perceptron (MLP)-based classification module (MCM) at the end of the method.
In summary, the main contributions are as follows: 
\begin{itemize}
	\item We propose a novel method for gait-based human emotion recognition, 
	learning the deep features from images encoded by skeleton joints and affective features.
	To the best of our knowledge, we are among the first to represent affective features as an image for gait-based human emotion recognition.
	\item Transformer-based complementarity module (TCM) is exploited to complement the information between skeleton joints and affective features effectively via capturing long range dependencies.
\end{itemize}

\vspace{-2mm}
\section{Methodology} \label{sec:method}

The overall framework of our proposed method consists of skeleton sequence encoding and two-stream emotion perception, as illustrated in \mbox{Fig. \ref{fig:framework}}.
Skeleton joints and affective features are first constructed via a skeleton sequence of gait, 
and encoded into skeleton joint image (SJI) and affective feature image (AFI), respectively.
Next, TNTC based on CNNs is modeled to extract hierarchical features whose complementary information are replenished cross TCMs. 
MCM is conducted at the end of the network to identify the emotion category.

\subsection{Skeleton sequence encoding}




\textbf{Skeleton joint image (SJI).} The motivation of encoding the skeleton sequence as an image is to take full advantage of CNN to compactly extract the local spatial-temporal features.
Inspired by \cite{Narayanan2020ProxEmo}, we encode the 3D coordinates of joints as three channel of the image. 
Specifically, a skeleton sequence is given as follows:
\vspace{-1mm}
\begin{equation}
S=\left\{ P^{t}  \in \mathbb{R}^{ N_{s} \times D_{s}} \big|  t=1, 2, \dots, T;  D_s=3 \right\},
\vspace{-1mm}
\end{equation}
where $P^{t}$ presents coordinates of skeleton joints in the $t$-th frame. 
$D_s$ denotes the dimension of coordinates, and the number of joints is denoted as $N_s$.

Then, we arrange $P^{t}$ according to the order of frames, and JFI is encoded as follows:
\vspace{-1mm}
\begin{equation}
\mathbf{M_{J}} = \bigg[ P^{1}, P^{2}, \dots, P^{T} \bigg],
\vspace{-1mm}
\end{equation}
where $\mathbf{M_{J}} \in \mathbb{R}^{T \times N_{s}\times D_{s} } $,  
and $[ \cdot ]$ is an operation to concatenate the coordinates of joints in the temporal dimension.

\begin{figure}[!t]\centering
	\includegraphics[width=0.9\linewidth]{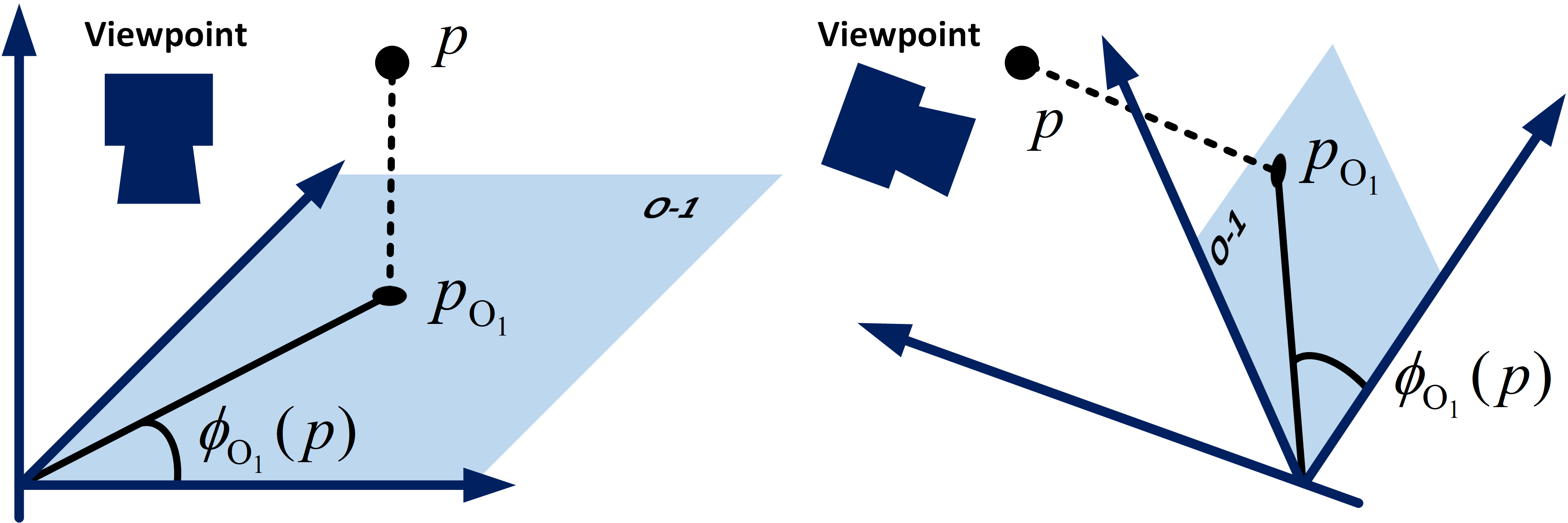}
	\vspace{-4mm}
	\caption{The visualization of invariant $\phi_{\mathbf{o}_i}({p})$ under arbitrary viewpoints.
	}
	\label{fig:point}
\end{figure}

\textbf{Affective feature image (AFI).} 
Besides the skeleton joints, affective features, such as posture and movement, convey the emotion information in the gaits \cite{xu2020emotion}
which are essential to involve the prediction for affective state of the subject.
Here, we merely consider the joint angles as the affective features, 
and discard other characteristics, such as distances and velocities of joints.
The reason is that the formulations of such characteristics can be approximated as special cases of a low-order combination
which are implicit in the convolution operations of joint stream.

To construct the reliable AFI,  
we utilize the projection angles of joints on three planes to avoid the inconsistency caused by various viewpoints.
As shown in \mbox{Fig. \ref{fig:point}}, the projection angles of joints on three planes are invariant with arbitrary viewpoints obviously.
Formally, the projection angle of the $n$-th joint $P^{t}_{n} \in \mathbb{R}^{D_{s}}$ on the projected planes can be computed as follows:
\vspace{-1mm}
\begin{equation}
Q^{t}_{n} = \bigg[\phi_{\mathbf{o}_1}(P^{t}_{n}), \phi_{\mathbf{o}_2}(P^{t}_{n}), ... ,\phi_{\mathbf{o}_i}(P^{t}_{n}) \bigg]
\vspace{-1mm}
\end{equation}
where $\phi_{\mathbf{o}_1}(P^{t}_{n})$ is a function to compute a projection angle of joint $P^{t}_{n}$ on plane $O$-$1$, 
and $\mathbf{o}_1$ is belonged to $\mathbf{O}$ which is a set of projection operators $\left\{ \mathbf{o}_1, \mathbf{o}_2, ... , \mathbf{o}_i | i=1, 2, \dots, D_s \right\}$. 
Since $P^{t}_{n}$ is represented via 3D coordinates, $D_s = 3$ and $\mathbf{O} = \left\{ \mathbf{o}_1, \mathbf{o}_2, \mathbf{o}_3  \right\}$ can be instantiated as followed:
\vspace{-1mm}
\begin{equation}
\mathbf{o}_1 = \big[e_1, e_2\big], \mathbf{o}_2 = \big[e_1, e_3\big], \mathbf{o}_3 = \big[e_2, e_3\big],
\vspace{-1mm}
\end{equation}
where $e_1, e_2, e_3 \in \mathbb{R}^{3}$ are unit vectors.
For a joint ${p}\in \mathbb{R}^{3}$, the function $\phi_{\mathbf{o}_i}(p)$ can be formulated as follows:
\vspace{-1mm}
\begin{equation}
\begin{aligned}
{p}_{\mathbf{o}_{i}} &= {p^{ \mathsf{T} }}  \mathbf{o}_{i}, \\
\phi_{\mathbf{o}_i}({p}) &= arctan \frac{p_{\mathbf{o}_{i}(2)}}{p_{\mathbf{o}_{i}(1)}+\epsilon},
\end{aligned}
\vspace{-1mm}
\end{equation}
where $p_{\mathbf{o}_{i}(1)}$ represents the 1-st value of the coordinate on the projected plane $O$-$i$, 
and $\epsilon$ is a small positive infinitesimal quantity to avoid the invalid denominator.

Finally, the angles of the skeleton joints $P^{t}$ in the $t$-th frame can be formulated as follows:
\vspace{-1mm}
\begin{equation}
Q^{t} = \bigg[Q^{t}_{1}, Q^{t}_{2}, \dots, Q^{t}_{N_s}\bigg],
\vspace{-1mm}
\end{equation}
where $Q^{t} \in \mathbb{R}^{ N_{s} \times D_{s}}$. AFI constructed via the projection angles can be denoted as follows:
\vspace{-1mm}
\begin{equation}
\mathbf{M_{A}} = \bigg[Q^{1}, Q^{2}, \dots, Q^{T} \bigg].
\vspace{-1mm}
\end{equation}
where $\mathbf{M_{A}} \in \mathbb{R}^{T \times N_{s}\times D_{s} } $, 
and $[ \cdot ]$ is an operation to concatenate the projection angles of joints in the temporal dimension.

\begin{figure}[!t]\centering
	\includegraphics[width=0.7\linewidth]{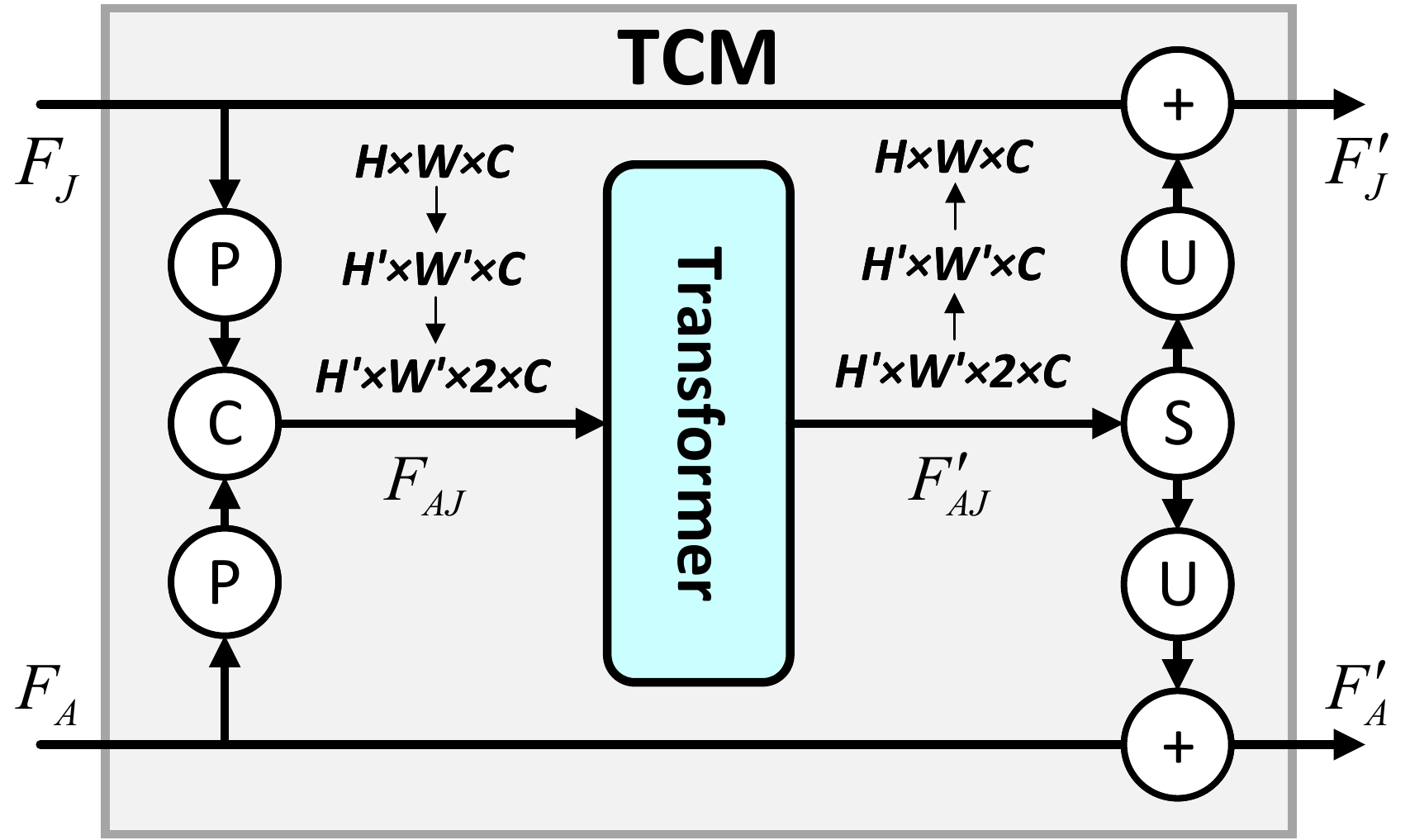}
	\caption{The details of TCM in TNTC. ``P'', ``C'', ``S'', and ``U'' represent average pooling operation, concatenation operation,
		splitting operation, and upsampling operation, respectively. Element-wise summation is denoted as ``$+$''.
	}
	\vspace{-1mm}
	\label{fig:tcm}
\end{figure}

\subsection{TNTC}

\textbf{Backbone.} The backbone of TNTC is ResNet-34 whose capability of feature representations has been proved in many vision tasks \cite{He2016Deep, Hu2020Efficient}.
We discard the fully connected layer of ResNet-34, and retain the four level residual blocks to extract hierarchical features, as shown in \mbox{Fig. \ref{fig:framework}}.
Assuming an input encoded map with a size of $224 \times 224$, 
the size and channel of output feature extracted via four level residual blocks are $7 \times 7$ and $512$, respectively.

\textbf{TCM.} 
After each residual block, TCM is modeled based on a vanilla transformer architecture \cite{dosovitskiy2020image} and a series of operations, as shown in \mbox{Fig. \ref{fig:tcm}}.
The transformer architecture takes as input a sequence consisting of discrete tokens, each represented by a feature vector. 
The long range dependencies between feature vectors are captured, 
and the feature vectors supplemented by positional encoding to incorporate positional inductive biases.
Specifically, given joints feature $F_{J}$ and affective feature $F_{A}$ extracted via residual blocks in a level,
the size of $F_{J}$ and $ F_{A}$ with $H \times W \times C$ are first reduced via average pooling operations to $H' \times W' \times C$,
where $H' = H/l$, $W' = W/l$, and scale parameter $l$ is set to different values to modify the features with a fixed size in each level, 
since processing features with transformer at high spatial resolutions is computationally expensive.
Then, two reduced features are concatenated to $F_{AJ} \in \mathbb{R}^{ H' \times W' \times 2 \times C}$ as input of transformer,
which is an encoder structure completely described in \cite{dosovitskiy2020image} and its implementation details are discussed in \mbox{Section \ref{sec:exp}}.
After capturing long range dependencies between $F_{J}$ and $F_{A}$ via transformer, 
$F'_{AJ}$ is arranged to $ H' \times W' \times 2 \times C$ and divided into two complementary information $F'^{c}_{J}$ and $F'^{c}_{A}$ with dimensions of $H' \times W' \times C$ for joint stream and affective stream, respectively.
Bilinear interpolation is utilized as upsampling operation to resize the complementary information to the same size of $F_{J}$ and $F_{A}$.
Finally, element-wise summation is leveraged to aggregate the complementary information and corresponding stream, respectively.

\textbf{MCM.} At the end of joint and affective streams, 
joint and affective features are reduced to a size of $1 \times 1 \times 512$ via average pooling,
and are combined by element-wise addition as the final feature vector with $512$-dimension.
The feature vector is fed to MLP with two hidden layers and softmax function to classify the emotion of the skeleton sequence.

\section{Experiments} \label{sec:exp}

\subsection{Experimental Setup}
\textbf{Dataset.} We evaluate the proposed method on Emotion-Gait \cite{Bhattacharya2020STEP} dataset, 
which consists of 2177 real gait sequences separately annotated into one of four emotion categories including happy, sad, angry, or neutral. 
The gait is defined as the 16 point pose model, and the steps of gait sequences are maintained via duplication to 240 which is the maximum length of gait sequence in the dataset.

\textbf{Evaluation protocol.} We employ 5-fold cross-validation to evaluate the proposed method, 
where the sample numbers of each category are divided in the same ratio among the folds.
Accuracy is adopted as the metric defined as follows:
\vspace{-1mm}
\begin{equation}
Accuracy = T / S,
\vspace{-1mm}
\end{equation}
where $T$ denotes to the number of successfully classified gait sequences with corresponding categories, and $S$ denotes the number of test samples.
The average accuracy of 5-fold cross-validation is recorded along its standard deviation.

\begin{table}[!tp]
	\begin{center}
		\caption{Comparison of our method with the state-of-the-art on Emotion-Gait. 
			The best result of accuracy is highlighted in \textbf{bold}.
		} \label{tab:sota}
		\vspace{1mm}
		\resizebox{0.8\columnwidth}{!}{
			\begin{tabular}{ccc}
				\toprule
				\multicolumn{2}{c}{Method}                & Accuracy \%  \\
				[-0mm]\midrule \\
				[-5.5mm]\midrule
				\rowcolor{gray!10}
				  & STEP \cite{Bhattacharya2020STEP}   & 77.65(0.87)             \\
				\rowcolor{gray!10}
				\multirow{-2}{*}{Graph-based}  & G-GCSN \cite{zhuang2020g}  & 80.31(0.92)             \\
				 & LSTM (Vanilla) \cite{randhavane2019identifying}    & 75.38(0.98)             \\
				\multirow{-2}{*}{Sequence-based} & TEW \cite{Bhattacharya2020Take}    & 81.89(0.69)             \\
				\rowcolor{gray!10}
				    & ProxEmo \cite{Narayanan2020ProxEmo}    & 80.33(0.85)             \\
				\rowcolor{gray!10}
				\multirow{-2}{*}{Image-based} & TNTC (\textbf{Ours}) & \textbf{85.97(0.75)} \\
				\bottomrule    
			\end{tabular}
		}
	\vspace{-4mm}
	\end{center}
\end{table}
		
\begin{table}[!tp]
	\begin{center}
		\caption{ Ablation analysis of our method on Emotion-Gait. 
		} \label{tab:abla}
		\vspace{1mm}
		\resizebox{0.8\columnwidth}{!}{
			\begin{tabular}{cccccc}
				\toprule
				\multicolumn{5}{c}{Method}                          & Accuracy \% \\
				[-0mm]\midrule \\
				[-5.5mm]\midrule
				\rowcolor{gray!10}
				\multicolumn{5}{c}{TNTC w/o TCMs (Baseline)}         & 81.53(0.88)             \\
				\multicolumn{5}{c}{Baseline w/o joint stream}       & 79.52(0.69)             \\
				\rowcolor{gray!10}
				\multicolumn{5}{c}{Baseline w/o affective stream}   & 80.71(0.71)             \\
				\midrule 
				Levels                              & 1 & 2 & 3 & 4 &              \\
				\rowcolor{gray!10}
				& \checkmark &   &   &   & 83.27(0.79)             \\
				\cellcolor{gray!10} & \checkmark & \checkmark &   &   & 84.09(0.91)             \\
				\rowcolor{gray!10}
				& \checkmark & \checkmark & \checkmark &   & 84.13(0.85)             \\
				\multirow{-4}{*}{\cellcolor{gray!10} Baseline with TCMs} & \checkmark & \checkmark & \checkmark & \checkmark & 85.97(0.75) 			\\
				\bottomrule 
			\end{tabular}
		}
		\vspace{-4mm}
	\end{center}
\end{table}

\textbf{Implementation Details.}
The experiments are conducted on a work station with an NVIDIA RTX 2080Ti GPU.
The proposed method is implemented based on PyTorch deep learning framework.
The main hyperparameters consist of the network and training stage.
For the hyperparameters of TCMs in the network, we stack 2 transformers and 4 attention heads for each TCM. 
The dimensions of feature embedding are 64, 128, 256, and 512 for corresponding TCMs, 
while the values of each level scale parameter $l$ are 8, 4, 2, and 1.
In the training stage, stochastic gradient descent (SGD) is used to optimize the learnable parameters with a momentum of 0.9 and a weight decay of 5e-4. 
The total epochs are 300 and initial learning rate is 1e-3, where the decay ratio is 0.1 every 75 epochs.
JFI and AFI are resized into $224 \times 224$ via bilinear interpolation, and the size of mini-batches is 64.

\subsection{Comparisons with the State-of-the-art}
We compare our method with 5 state-of-the-art methods lately reported on Emotion-Gait including sequence-based \cite{Bhattacharya2020Take,randhavane2019identifying}, 
graph-based \cite{Bhattacharya2020STEP, zhuang2020g}, and image-based \cite{Narayanan2020ProxEmo} methods.  
To come up with a fair comparison, we reproduce these methods by public codes and report the experimental results with the same evaluation protocol, 
as listed in \mbox{Tab. \ref{tab:sota}}.
We can observe that the proposed method achieves a superior performance than all the other methods.

\subsection{Ablation studies}

\textbf{Effectiveness of two-stream architecture.} 
To clarify the effectiveness of two-stream architecture, 
we separately train joint stream and affective stream as two independent networks.
Meanwhile, we construct TNTC without TCMs as a baseline network.
As shown in \mbox{Tab. \ref{tab:abla}}, the performances of individual streams are obviously lower than the baseline.
The results demonstrate the effectiveness of two-stream architecture 
which implies the complementary relationship between joint and affective features.

\begin{figure}[!t]\centering
	\includegraphics[width=\linewidth]{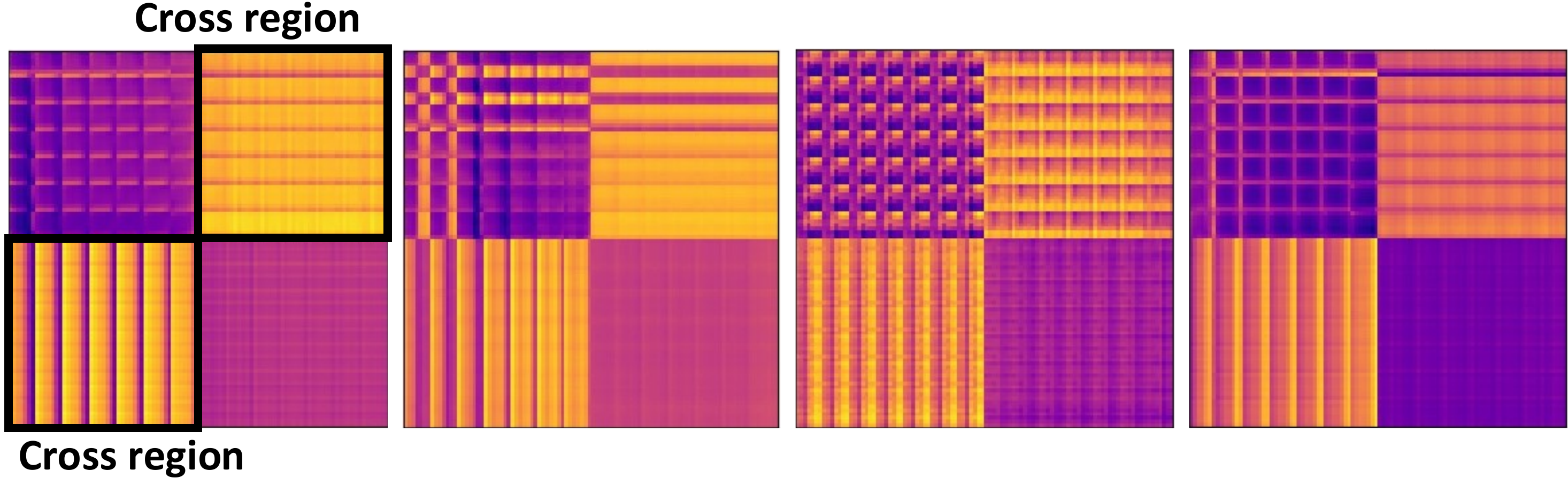}
	\caption{The visual interpretation of complementary information. 
		The average attention maps in TCMs of different instances are ordered by four categories from happy, sad, angry to neutral.
		The cross regions (\textbf{bold} squares) are highlighted obviously indicating that 
		TCM can capture promising complementary information between two streams to improve the performance of TNTC.
	}
	\vspace{-2mm}
	\label{fig:trans_vis}
\end{figure}

\textbf{Effectiveness of TCMs.} 
To confirm the effectiveness of TCMs, we gradually insert TCM at each level based on the baseline,
and report the performances of the networks in \mbox{Tab. \ref{tab:abla}}, respectively.
It can be observed that the accuracy of the baseline is improved as the number of TCM increases.
Furthermore, we plot the attention maps of TCMs to reveal the capability of representing complementary information.
As shown in \mbox{Fig. \ref{fig:trans_vis}}, the high scores of the attention maps focus on the cross regions between skeleton joints and affective features,
which interpret visually the complementary information between two streams represented via TCMs.

\section{Conclusion} \label{sec:con}
In this paper, we propose a novel method for gait-based human emotion recognition,
modeled via a two-stream network with transformer-based complementarity (TNTC).
Skeleton joints and affective features of gait sequence are encoded into images as the inputs of the two-stream architecture. 
Meanwhile, the importance of complementary information between two streams is revealed,
which can be represented effectively via the proposed transformer-based complementarity module (TCM).
Experimental results demonstrate that the proposed method achieves the superior performance over state-of-the-art methods on the latest gait-based emotion dataset.

\bibliographystyle{IEEEbib}
\bibliography{emotion_icassp}

\end{document}